\pdfoutput=1
\documentclass{article}

\PassOptionsToPackage{numbers, compress}{natbib}

\usepackage[preprint]{neurips_2026}

\usepackage[utf8]{inputenc} 
\usepackage[T1]{fontenc}    
\usepackage{hyperref}       
\usepackage{url}            
\usepackage{booktabs}       
\usepackage{amsfonts}       
\usepackage{nicefrac}       
\usepackage{microtype}      
\usepackage{xcolor}         
\usepackage[most]{tcolorbox}
\usepackage{subcaption}
\usepackage{wrapfig}
\usepackage{mathtools}
\usepackage{xcolor}
\usepackage{tabularx}
\usepackage{booktabs}

\title{SinkRec: Mitigating Semantic State Sink in Long Sequence Recommendation with Memory-Conditioned Gated Delta Networks}

\author{%
Zhuang~Zhuang\textsuperscript{1},~~
Zhipeng~Wei\textsuperscript{1},~~
Ji~Dai\textsuperscript{2},~~
Jie~Chen\textsuperscript{1}\thanks{Corresponding author.},~~
Fei~Pan\textsuperscript{1}\footnotemark[1],~~
Peng~Jiang\textsuperscript{1},~~
Kun~Gai\textsuperscript{3} \\
\textsuperscript{1}Kuaishou Technology, Beijing, China\\
\textsuperscript{2}Beijing University of Posts and Telecommunications,
Beijing, China\\
\textsuperscript{3}Independent Researcher\\
\texttt{\{zhuangzhuang,weizhipeng,chenjie20,panfei05,jiangpeng\}@kuaishou.com}\\
\texttt{daiji@bupt.edu.cn}, \texttt{gai.kun@qq.com}\\
}

\begin{document}

\maketitle

\begin{abstract}
Linear attention provides an efficient backbone for long-sequence recommendation by avoiding the quadratic cost of standard Transformers, but its compressed recurrent state can be dominated by repetitive behavior patterns. 
We identify this phenomenon as \textit{semantic state sink}, where recurring semantics over-occupy the recurrent state and bias subsequent readouts. 
To mitigate semantic state sink, we propose \textbf{SinkRec}, a hybrid memory-transition looped architecture that decouples collaborative behavioral pattern storage from dynamic transition modeling. 
SinkRec externalizes recurring local patterns into a learnable conditional memory through residual vector quantization, reinjects the retrieved codes, and exposes memory key-value pairs to the attention block. 
It further introduces Temporal-Aware State-Relation Differential Gated DeltaNet (TDGD), which uses memory to purify recurrent writing and reading by suppressing memory-covered updates and removing memory-aligned readout responses. 
This design turns recurring semantics from state-competing signals into memory-retrievable patterns, allowing the recurrent state to focus on dynamic transitions and alleviating semantic state sink with linear-time efficiency.
Experiments on public and industrial datasets demonstrate the effectiveness and efficiency of SinkRec. 
\end{abstract}

\section{Introduction}
Sequential recommendation is fundamental to personalized services, such as streaming
media and e-commerce platforms, as it models users' historical behavior sequences to capture personalized interests and deliver relevant content to users~\cite{zhuang2025mgstdn, zhuang2026think2go, yang2026sarm}. 
As such, sequential modeling has become a fundamental approach for capturing evolving user interests. Early recommendation architectures adopted temporal models such as Markov chains, RNNs, and Transformers, but were mostly applied to short sequences (lengths of $10^2$--$10^3$). In contrast, full long sequences (length>$10^3$) reveal long-term preferences, recurring interests, and delayed dependencies, improving recommendation accuracy and helping mitigate the information cocoon effect. This makes scalable long-sequence modeling a critical step toward more comprehensive and less myopic user preference modeling.

Existing long-sequence recommendation methods often trade off efficiency and completeness. 
Search-based methods~\cite{zhou2018deep, zhou2019deep, pi2020search, chang2023twin} reduce computation by retrieving partial histories, but introduce two-stage serving complexity and incomplete interest estimation. 
End-to-end models~\cite{zhai2024actions} retain richer historical signals but incur rapidly increasing computation. 
This motivates efficient backbones such as linear attention, which scale to long histories while preserving sequence perception.

Although linear attention offers an efficient backbone for long-sequence recommendation, its recurrent-state formulation creates a new bottleneck for long-history modeling.
It compresses the whole history into a finite state matrix, where each behavior writes its key-value information into the state used for future prediction. 
This avoids quadratic attention cost, but also couples recurring semantic storage with dynamic transition modeling.
In long histories, recurring behaviors provide useful preference regularities, whereas sparse transitions reflect changes in current intent. 
When both share the same compressed state, repetitive semantics can be repeatedly reinforced and interfere with transition signals needed for the current prediction. 
This motivates a central question:

\begin{tcolorbox}[
    colback=gray!10,
    colframe=gray!10,
    boxrule=0pt,
    arc=1pt,
    left=6pt,
    right=6pt,
    top=4pt,
    bottom=4pt
]
\textit{How can recurrent linear attention exploit long user histories without allowing repetitive semantics to dominate the recurrent state?}
\end{tcolorbox}

\begin{wrapfigure}[24]{r}{0.56\linewidth}
    \centering
    \vspace{-6pt}
    \includegraphics[width=\linewidth]{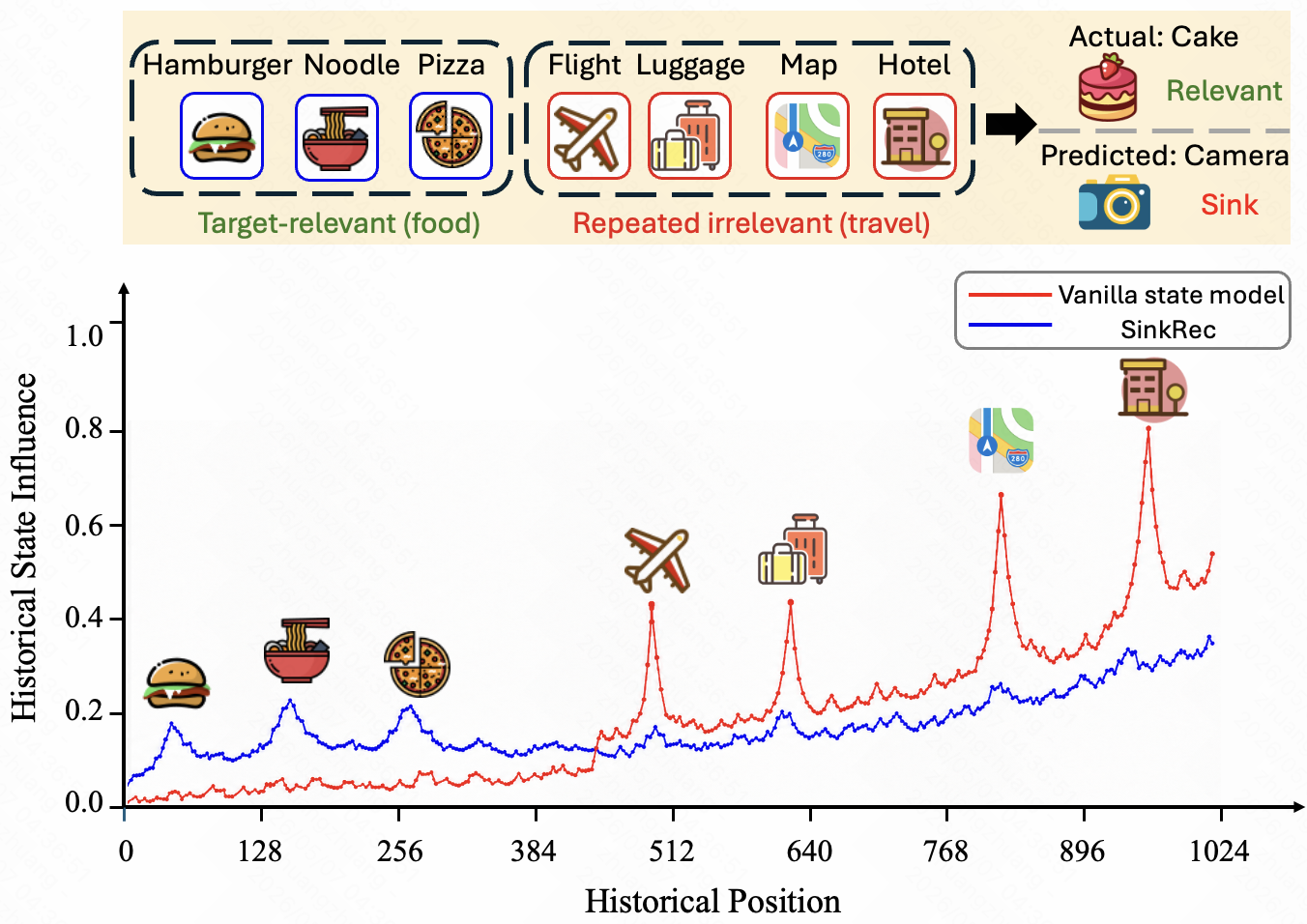}
    \vspace{-8pt}
    \caption{
    Motivating example of repetitive semantic state sink.
    The y-axis measures each past behavior's contribution to the current prediction through the recurrent state; higher values indicate stronger occupation of the predictive state.
    Vanilla recurrent states are dominated by repeated irrelevant semantics, while SinkRec suppresses this sink and preserves target-relevant signals.
    }
    \label{fig:intro}
    \vspace{-8pt}
\end{wrapfigure}

To answer this question, we analyze how Gated DeltaNet-style recurrent attention carries historical behaviors into the current prediction. 
Figure~\ref{fig:intro} illustrates a case where early food-related behaviors match the actual next item, while recent repetitive travel-related behaviors are target-irrelevant but receive disproportionately high historical influence scores~(detailed in Appendix~\ref{app:state_influence}) in vanilla Gated DeltaNet.
This indicates that the current prediction is dominated by sink-like travel semantics rather than target-relevant food signals.
This exposes a key challenge: the compressed recurrent state is required to serve simultaneously as semantic memory and transition operator, making it vulnerable to semantic state sink when repetitive patterns are over-retained and dominate subsequent readouts.

To address the challenges introduced by the semantic state sink phenomenon, we propose \textbf{SinkRec}, a memory-transition decoupled framework for efficient long-sequence recommendation through a looped hybrid architecture. 
The key insight is to separate collaborative semantic storage from dynamic transition modeling: recurring local behavior patterns are externalized into conditional memory, while the recurrent state is reserved for memory-unexplained transitions. 
Specifically, SinkRec consists of two complementary components: 
(i) \textit{Conditional Memory Module} compresses local behavior windows into learnable residual vector-quantized (VQ) codes, reinjects the retrieved codes into the sequence, and exposes memory key--value pairs to the downstream attention block. 
(ii) \textit{Temporal-Aware State-Relation Differential Gated DeltaNet} (TDGD) performs time-aware recurrent modeling and uses the memory pairs to purify state writing and reading: memory-covered updates are suppressed before being written, and memory-aligned responses are removed during readout. 
In this way, SinkRec leverages shared behavioral patterns without repeatedly accumulating them in the recurrent state, mitigating semantic state sink while preserving the efficiency of recurrent linear attention. Moreover, parameter sharing across hybrid architecture blocks keeps SinkRec compact while maintaining strong recommendation performance.

Our contributions are summarized as follows:

\begin{itemize}

    \item We identify the \emph{semantic state sink} phenomenon in Gated DeltaNet-style long-sequence recommendation, where semantically repetitive patterns can over-occupy the compressed recurrent state and bias subsequent readouts.
    \item We propose \textbf{SinkRec}, a memory-transition decoupled framework that externalizes recurring local patterns into conditional memory and uses Temporal-Aware State-Relation Differential Gated DeltaNet (TDGD) to purify recurrent writing and reading from memory-covered semantics.
    \item We conduct extensive experiments on two public datasets and one industrial dataset. The results show that SinkRec consistently outperforms strong baselines with fewer parameters, demonstrating both effectiveness and efficiency for long-sequence recommendation.
\end{itemize}

\section{Related Work}
\label{related_work}
\textbf{Long Sequential Recommendation Architectures.}
Scaling long-sequence user interaction histories has been increasingly explored as an effective means of improving recommender model performance~\cite{lai2026unleashing}. Existing methods commonly rely on attention mechanisms to capture complex user transition patterns and facilitate personalized recommendation, while continuously seeking to exploit the information gain brought by long histories under manageable computational costs. Early studies such as DIN~\cite{zhou2018deep} and SIM~\cite{pi2020search} adopt search-based mechanisms to retrieve valuable subsets from historical interaction sequences, inspiring subsequent methods such as VISTA~\cite{chen2025massive}, which caches user histories into a few hundred compact tokens to jointly support prediction. Later, HSTU~\cite{zhai2024actions} reframes recommendation as a sequential transduction task and customizes attention mechanisms for large-scale, non-stationary recommendation data. More recently, LONGER\cite{chai2025longer}, HiSAC~\cite{yuan2026hisac}, and GEMs~\cite{zhou2026gems} further exploit Transformer-based architectures to model relevance-driven interactions.
However, the quadratic complexity of Transformer-based attention limits its scalability to ultra-long user histories, motivating the emergence of linear-complexity alternatives. Methods such as RankMixer~\cite{zhu2025rankmixer} and UniMixer~\cite{ha2026unimixer} enhance the expressive capacity of mixing modules to improve interaction modeling. BlossomRec~\cite{ma2026blossomrec} employs sparse attention to capture both long- and short-term user interests, while FuXi-Linear~\cite{ye2026fuxi} integrates linear attention with temporal features for efficient long-sequence modeling. Nevertheless, these methods inherently lack a knowledge lookup mechanism, forcing them to approximate transition relations purely through continuous computation rather than retrieving collaborative semantic patterns from external or structured memory.

\textbf{Memory-Augmented Model Scaling.}
Recent studies~\cite{behrouz2026memory,cheng2026conditional} have shown that incorporating memory modules into model backbones can enhance model capacity and improve scaling behavior. In the field of large language models~\cite{zhao2023survey}, LongMem~\cite{wang2023augmenting} introduces a long-term memory and retrieval mechanism to efficiently leverage ultra-long contexts. UltraMem~\cite{huang2024ultra} and UltraMemV2~\cite{huang2025ultramemv2} replace sparsely activated experts with efficient memory layers, thereby reducing memory-access overhead. Engram~\cite{cheng2026conditional} further proposes conditional memory as a complementary sparse dimension for scaling the capacity of LLMs. In recommender systems, early work such as MIMN~\cite{pi2019practice} captures evolving user interests for long-sequence modeling. More recently, MSN~\cite{wu2026msn} and related methods~\cite{lu2025large,liu2024dynamic,chen2026recurrent} retrieve personalized representations from large parameterized memories and aggregate them into downstream feature-interaction modules. However, existing recommender models primarily exploit the retrieval capability of memory modules, while largely overlooking their potential to complement and collaborate with sequential modeling modules within a unified architecture.

\section{Preliminaries}
\subsection{Gated Delta Networks}
Linear Transformers improve efficiency over standard Transformers, but their reduced contextual interaction often limits performance on long-context tasks. 
Gated DeltaNet addresses this issue by extending DeltaNet with adaptive memory-control gates and a delta-update rule.
Given the query, key, and value vectors $\mathbf{q}_t$, $\mathbf{k}_t$, and $\mathbf{v}_t$ at step $t$, Gated DeltaNet maintains a key-addressed recurrent state:
\begin{equation}
\mathbf{S}_t
=
\alpha_t
\mathbf{S}_{t-1}
\left(
\mathbf{I}
-
\beta_t\mathbf{k}_t\mathbf{k}_t^{\top}
\right)
+
\beta_t\mathbf{v}_t\mathbf{k}_t^{\top},
\qquad
\mathbf{o}_t
=
\mathbf{S}_t\mathbf{q}_t ,
\end{equation}
where $\mathbf{S}_t\in\mathbb{R}^{d_v\times d_k}$ is the recurrent state, $\alpha_t$ controls state retention, and $\beta_t$ controls the delta-update strength.
With the cumulative decay $\gamma_j=\prod_{i=1}^{j}\alpha_i$, the recurrence admits an attention-like form:
\begin{equation}
\mathbf{o}_t
=
\sum_{i=1}^{t}
\mathbf{v}_i
\left(
\frac{\gamma_t}{\gamma_i}
\mathbf{k}_i^{\top}\mathbf{q}_t
\right),
\qquad
\mathbf{O}
=
\left(
\mathbf{Q}\mathbf{K}^{\top}
\odot
\mathbf{\Gamma}
\right)
\mathbf{V},
\end{equation}
where $\mathbf{\Gamma}\in\mathbb{R}^{L\times L}$ is a causal decay mask that assigns decay-aware weights to visible historical positions and masks future positions.

For efficient training, Gated DeltaNet adopts a chunkwise parallel formulation. 
The transition matrix $\mathbf{I}-\beta_t\mathbf{k}_t\mathbf{k}_t^{\top}$ can be viewed as a generalized Householder transformation, whose cumulative product is represented with a WY-style factorization under a partially expanded recurrence, enabling efficient parallel computation.

\subsection{Semantic State Sink}
To understand why Gated DeltaNet-style recurrent attention may underutilize long user histories, we analyze the historical influence on the current prediction in Figure~\ref{fig:intro}. 
It shows that early target-relevant behaviors can still provide useful long-range signals, but repeated target-irrelevant semantics in recent history may produce increasingly dominant state responses in vanilla Gated DeltaNet. 
As a result, the recurrent state is biased toward a sink-like semantic direction, causing the prediction to over-rely on recurring irrelevant patterns while weakening the influence of target-relevant long-range behaviors.

\begin{tcolorbox}[
    colback=gray!10,
    colframe=gray!10,
    boxrule=0pt,
    arc=1pt,
    left=6pt,
    right=6pt,
    top=4pt,
    bottom=4pt
]
\textbf{Phenomenon 1. (Semantic State Sink)} \\
Semantic state sink refers to the over-retention of semantically repetitive patterns in the recurrent state, where a few semantic directions dominate the state readout and bias subsequent predictions.
\end{tcolorbox}

This phenomenon reveals both the opportunity and challenge of long-sequence recommendation. 
Long histories contain collaborative behavioral patterns beyond short-term contexts, but directly accumulating them in the recurrent state couples semantic storage with transition modeling. 
As repetitive patterns are repeatedly written into similar state directions, they may form sink-like semantic directions that dominate subsequent readouts and suppress memory-unexplained transition signals. 
Thus, collaborative patterns should be externalized into memory, while recurrent computation should focus on memory-unexplained transitions. 
SinkRec follows this principle by combining conditional memory with TDGD to leverage shared behavioral patterns and purify recurrent writing and reading from memory-covered semantics.

\begin{figure*}[!t]
\centering	
\includegraphics[width=\textwidth]{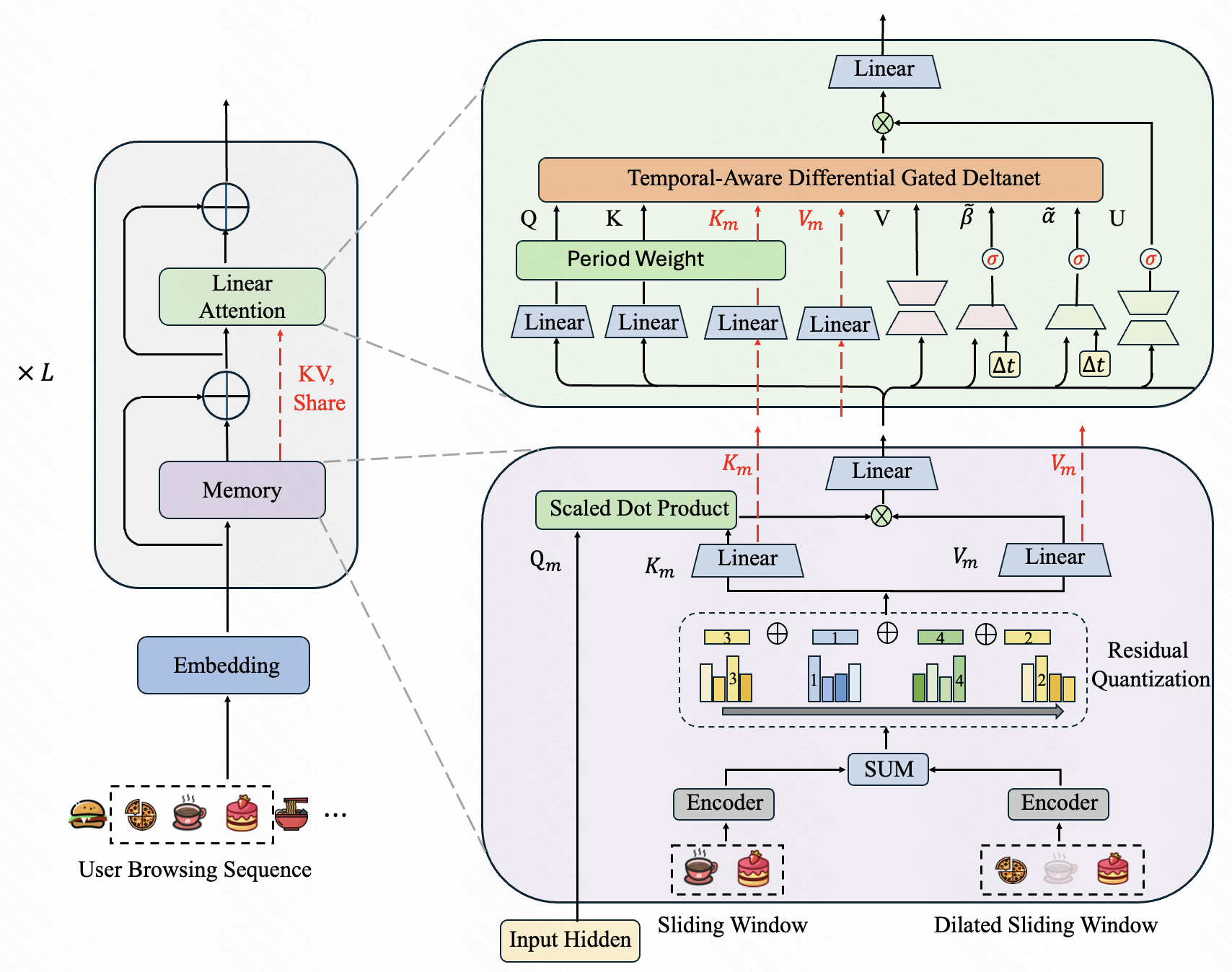}
\caption{The Architecture of SinkRec for Long Sequence Recommendation. 
SinkRec consists of $L$ stacked hybrid architecture blocks, each composed of a conditional memory module and a Gated DeltaNet-style linear attention module. 
The conditional memory module encodes contiguous and dilated behavior windows into residual-quantized memory representations, using the dilated window to enlarge the historical receptive field and sharing the resulting memory key--value pairs with the linear attention module.
The linear attention module then performs temporal-aware Gated DeltaNet updates with memory-conditioned writing and state-relation differential readout.
}
\label{fig:architecture}
\end{figure*}

\section{Method}
\label{method}

\subsection{Hybrid Model Architecture Overview}
\label{sec:architecture_overview}
As shown in Figure~\ref{fig:architecture}, SinkRec adopts a looped hybrid architecture that combines conditional memory with TDGD to separate collaborative pattern storage from dynamic transition modeling~\cite{gao2026hysparse}.
Section~\ref{sec:conditional_memory} provides the memory side, which preserves recurring local behavior patterns outside the recurrent state. 
Section~\ref{sec:time_aware_attention} provides the dynamic computation side, which purifies recurrent writing and reading by suppressing memory-covered updates and removing memory-aligned readout responses. 
Together, they allow SinkRec to reuse recurring patterns while preventing them from repeatedly dominating the recurrent state. We further provide a theoretical analysis of semantic state sink in Appendix~\ref{app:recurrent_attention_sink}, showing how repetitive semantic patterns can dominate recurrent state readouts.

\subsection{Conditional Memory Module}
\label{sec:conditional_memory}
Long sequence recommendation involves recurring interests that appear as collaborative behavior patterns. 
Motivated by this, SinkRec reduces redundant state encoding by externalizing them into a learnable conditional memory, where compressed behavior windows are residual-quantized, reinjected, and exposed as memory key--value pairs to TDGD.

\paragraph{Behavior Window Encoding.}
Given the dense behavioral representation $\mathbf{x}_{1:T}$, SinkRec constructs two local windows for each position $t$: a contiguous window and a dilated window. 
The contiguous window captures short-range behavioral transitions, while the dilated window enlarges the receptive field over earlier interactions:
\begin{equation}
\mathbf{W}_{t}^{\mathrm{con}}
=
[\mathbf{x}_{t-W+1},\ldots,\mathbf{x}_{t}],
\qquad
\mathbf{W}_{t}^{\mathrm{dil}}
=
[\mathbf{x}_{t-(W-1)s},\ldots,\mathbf{x}_{t-s},\mathbf{x}_{t}],
\end{equation}
where $W$ is the window size and $s$ is the dilation stride. 
The consecutive and dilated windows are encoded separately and then summed as the compressed local behavior representation, i.e., $\mathbf{h}_t=\mathrm{Enc}_{\mathrm{con}}(\mathbf{W}_{t}^{\mathrm{con}})+\mathrm{Enc}_{\mathrm{dil}}(\mathbf{W}_{t}^{\mathrm{dil}})$, which is used for subsequent memory quantization.

\paragraph{Learnable Residual Vector Quantization.}
SinkRec discretizes $\mathbf{h}_t$ using a multi-level learnable residual VQ codebook. 
Starting from $\mathbf{r}_t^{0}=\mathbf{h}_t$, the $\ell$-th quantization level selects the nearest codeword and updates the residual as:
\begin{equation}
c_t^{\ell}
=
\arg\min_{k}
\left\|
\mathbf{r}_t^{\ell}
-
\mathbf{e}_{k}^{\ell}
\right\|_2^2,
\quad
\mathbf{q}_t^{\ell}
=
\mathbf{e}_{c_t^{\ell}}^{\ell},
\quad
\mathbf{r}_t^{\ell+1}
=
\mathbf{r}_t^{\ell}
-
\mathbf{q}_t^{\ell},
\quad
\mathbf{z}_t^q
=
\sum_{\ell=0}^{L-1}
\mathbf{q}_t^{\ell},
\end{equation}
where $\mathbf{e}_{k}^{\ell}$ denotes the $k$-th learnable codeword in the $\ell$-th codebook. 
Rather than relying on static semantic clusters, the codebooks are optimized end-to-end to learn recurring behavioral patterns.

\paragraph{Memory Output.}
The quantized representation $\mathbf{z}_t^q$ produces two types of outputs. 
First, it generates a code sequence that is injected back into the main sequence representation:
\begin{equation}
\mathbf{k}_t^{q}
=
\mathbf{W}_{k}^{q}\mathbf{z}_t^q,
\qquad
\mathbf{v}_t^{q}
=
\mathbf{W}_{v}^{q}\mathbf{z}_t^q,
\qquad
g_t^{q}
=
\sigma
\left(
\frac{
\mathrm{Norm}(\mathbf{k}_t^{q})^{\top}
\mathrm{Norm}(\mathbf{x}_t)
}{
\sqrt{d}
}
\right),
\end{equation}
\begin{equation}
\mathbf{c}_t
=
g_t^{q}\mathbf{v}_t^{q},
\qquad
\mathbf{x}_t = \mathbf{x}_t+\mathbf{c}_t ,
\end{equation}
where $\mathbf{c}_t$ denotes the retrieved code representation injected through a residual connection. 
The projected quantized representations are also exposed to TDGD as memory pairs, $\mathbf{k}_{\mathrm{mem},t}=\mathbf{k}_t^{q}$ and $\mathbf{v}_{\mathrm{mem},t}=\mathbf{v}_t^{q}$, providing semantic addresses and collaborative behavioral content for memory-conditioned writing and differential readout.

\paragraph{Training Objective.}
The memory module is optimized with residual VQ and a reconstruction loss:
\begin{equation}
\mathcal{L}_{\mathrm{mem}}
=
\sum_{\ell=0}^{L-1}
\left(
\|\mathrm{sg}[\mathbf{h}_t]-\mathbf{q}_t^{\ell}\|_2^2
+
\beta_c
\|\mathbf{h}_t-\mathrm{sg}[\mathbf{q}_t^{\ell}]\|_2^2
\right)
+
\lambda_{\mathrm{rec}}
\left\|
\mathrm{Dec}(\mathbf{z}_t^q)-\mathbf{Y}_t
\right\|_2^2 ,
\end{equation}
where $\mathrm{sg}[\cdot]$ denotes the stop-gradient operator, $\beta_c$ controls the commitment strength, $\mathbf{Y}_t$ is the subsequent behavior window used as reconstruction supervision for the next-behavior prediction context. 
The final objective is $\mathcal{L}=\mathcal{L}_{\mathrm{pred}}+\lambda_{\mathrm{vq}}\mathcal{L}_{\mathrm{mem}}$.

\subsection{Temporal-Aware State-Relation Differential Gated DeltaNet}
\label{sec:time_aware_attention}
We identify \emph{semantic state sink} in recurrent linear attention, where recurring semantic interests repeatedly accumulate in the recurrent state and dominate future readouts. 
TDGD mitigates this issue by using external semantic memory to regulate both state writing and state reading.

\paragraph{Temporal-aware Gated DeltaNet.}
Given the hidden representation $\mathbf{x}_t$, TDGD obtains the block-level gate, value, query, and key vectors through a joint SiLU activated projection:
\begin{equation}
\left[
\mathbf{u}_t,\mathbf{v}_t,\mathbf{q}_t,\mathbf{k}_t
\right]
=
\operatorname{Split}
\left(
\operatorname{SiLU}
\left(
\mathbf{x}_t\mathbf{W}_{uvqk}
\right)
\right),
\end{equation}
where $\mathbf{q}_t$ and $\mathbf{k}_t$ address the recurrent state, $\mathbf{v}_t$ provides the written content, and $\mathbf{u}_t$ serves as a block-level output gate that modulates the recurrent readout before the feed-forward transformation.
Following the gated delta rule, vanilla Gated DeltaNet maintains a recurrent key--value state as:
\begin{equation}
\mathbf{S}_t
=
\alpha_t\mathbf{S}_{t-1}
\left(
\mathbf{I}-\beta_t\mathbf{k}_t\mathbf{k}_t^{\top}
\right)
+
\beta_t\mathbf{v}_t\mathbf{k}_t^{\top},
\qquad
\mathbf{r}_t=\mathbf{S}_t\mathbf{q}_t ,
\end{equation}
where $\alpha_t$ controls state retention and $\beta_t$ controls write strength. The readout is further normalized, projected, and modulated by $\mathbf{u}_t$ before being passed to the downstream feed-forward layer.

However, user behaviors in long sequences are highly time-dependent. 
TDGD therefore injects temporal signals into both query-key addressing and recurrent gating. 
Following the multi-granularity temporal encoding in FuXi-Linear~\cite{ye2026fuxi}, we define temporal periods $P_j=B^{b_0+j}$ and represent each timestamp $\tau$ with periodic phases 
$\{\sin(2\pi\tau/P_j),\cos(2\pi\tau/P_j)\}_{j=0}^{m-1}$, denoted as $\boldsymbol{\phi}(\tau)$. 
The key is conditioned on the current interaction time, while the query is conditioned on the target time:
\begin{equation}
\tilde{\mathbf{k}}_t
=
\operatorname{Norm}
\left(
\mathbf{k}_t+\mathbf{W}_{k}^{\tau}\boldsymbol{\phi}(\tau_t)
\right),
\qquad
\tilde{\mathbf{q}}_t
=
\operatorname{Norm}
\left(
\mathbf{q}_t+\mathbf{W}_{q}^{\tau}\boldsymbol{\phi}(\tau_{t+1})
\right).
\end{equation}

The time interval and periodic feature jointly modulate state retention:
\begin{equation}
\omega_t
=
\underbrace{
\exp\left(
-\lambda_{\Delta}
\log\left(1+\frac{\Delta t_t}{\tau_{\Delta}}\right)
\right)
}_{\text{interval-aware temporal decay}}
\cdot
\underbrace{
\sigma\left(\mathbf{w}_{\phi}^{\top}\boldsymbol{\phi}(\tau_t)\right)
}_{\text{periodic preference}},
\end{equation}
where $\Delta t_t$ denotes the interaction interval, and $\tau_{\Delta}$ and
$\lambda_{\Delta}$ control the scale and strength of temporal decay. The final
retention and write gates are:
\begin{equation}
\alpha_t
=
\sigma(\mathbf{x}_t\mathbf{W}_{\alpha})\,\omega_t,
\qquad
\beta_t
=
\sigma
\left(
\mathbf{x}_t\mathbf{W}_{\beta}
+
\mathbf{W}_{\Delta}\Delta t_t
\right).
\end{equation}

With these temporal components, TDGD writes the recurrent update in innovation
form:
\begin{equation}
\bar{\mathbf{S}}_{t-1}
=
\alpha_t\mathbf{S}_{t-1},
\qquad
\boldsymbol{\delta}_t
=
\mathbf{v}_t
-
\bar{\mathbf{S}}_{t-1}\tilde{\mathbf{k}}_t,
\qquad
\mathbf{S}_t
=
\bar{\mathbf{S}}_{t-1}
+
\beta_t
\boldsymbol{\delta}_t
\tilde{\mathbf{k}}_t^{\top}.
\end{equation}
The raw recurrent readout is then obtained as $\mathbf{r}_t=\mathbf{S}_t\tilde{\mathbf{q}}_t$.
This enables TDGD to preserve temporally relevant states while adapting updates to time-aware transitions.

\paragraph{Memory-conditioned write gate.}
The memory module retrieves codebook-based key--value representations
$\mathbf{k}_{\mathrm{mem},t}$ and $\mathbf{v}_{\mathrm{mem},t}$. We first map
them into the GDN write space through SiLU-activated projections:
\begin{equation}
\mathbf{k}_t^m
=
\operatorname{SiLU}
\left(
\mathbf{k}_{\mathrm{mem},t}\mathbf{W}_{km}
\right),
\qquad
\mathbf{v}_t^m
=
\operatorname{SiLU}
\left(
\mathbf{v}_{\mathrm{mem},t}\mathbf{W}_{vm}
\right).
\end{equation}

To determine whether the current state update is already covered by the
retrieved memory, we compare the current write pattern with the memory-induced
write pattern:
\begin{equation}
\mathbf{P}_t
=
\mathbf{v}_t\tilde{\mathbf{k}}_t^{\top},
\qquad
\mathbf{M}_t
=
\mathbf{v}_t^m(\mathbf{k}_t^m)^{\top},
\end{equation}
where $\mathbf{P}_t$ denotes the current key--value write, and $\mathbf{M}_t$
denotes the memory-side write. We compute their write magnitudes and overlap as:
\begin{equation}
a_t^w
=
\|\mathbf{P}_t\|_F^2,
\qquad
b_t^w
=
\|\mathbf{M}_t\|_F^2,
\qquad
s_t^w
=
(\mathbf{v}_t^{\top}\mathbf{v}_t^m)
(\tilde{\mathbf{k}}_t^{\top}\mathbf{k}_t^m),
\end{equation}
where $a_t^w$ and $b_t^w$ measure the strengths of the two writes, while
$s_t^w$ measures their key--value overlap. The memory-unexplained write energy
is:
\begin{equation}
e_t^w
=
a_t^w+\lambda_w^2 b_t^w-2\lambda_w s_t^w
=
\left\|
\mathbf{P}_t-\lambda_w\mathbf{M}_t
\right\|_F^2 ,
\end{equation}
where $\lambda_w$ is a learnable scale for calibrating the memory-side write.

However, the gate should depend on both the residual energy and the scale of
the current and memory-side writes. Therefore, TDGD converts the overlap,
residual energy, and write magnitudes into a memory coverage score:
\begin{equation}
c_t^w
=
\sigma
\left(
\mathbf{W}_c^{\top}
\Phi_w(a_t^w,b_t^w,s_t^w,e_t^w)
+
b_w
\right),
\end{equation}
where $\Phi_w(\cdot)$ is a compact coverage feature that summarizes the write
overlap, residual energy, and write magnitudes. The write gate is then defined as:
\begin{equation}
g_t^w
=
1-\eta_w c_t^w,
\qquad
\bar{\mathbf{v}}_t
=
g_t^w\mathbf{v}_t .
\end{equation}
Thus, when the current write is largely covered by memory, $c_t^w$ becomes
larger and the effective value $\bar{\mathbf{v}}_t$ is suppressed; otherwise,
the update is preserved. Finally, TDGD updates the recurrent state using the
memory-filtered value and the time-aware key:
\begin{equation}
\bar{\mathbf{S}}_{t-1}
=
\alpha_t\mathbf{S}_{t-1},
\qquad
\mathbf{S}_t
=
\bar{\mathbf{S}}_{t-1}
+
\beta_t
\left(
\bar{\mathbf{v}}_t
-
\bar{\mathbf{S}}_{t-1}\tilde{\mathbf{k}}_t
\right)
\tilde{\mathbf{k}}_t^{\top}.
\end{equation}
This write gate reduces redundant semantic accumulation in the recurrent state
while retaining memory-unexplained transition signals.

\paragraph{State-relation differential readout.}
Although the write gate limits future memory-covered updates, the recurrent
state may already contain accumulated semantic components~\cite{pu2025linear}. We further apply a
memory-guided differential readout to suppress these residual memory-aligned
responses at query time. Given the normalized query and memory-key directions,
denoted by $\bar{\mathbf{q}}_t$ and $\bar{\mathbf{k}}_t^m$, we compute two
readouts from the recurrent state:
\begin{equation}
\mathbf{r}_t^q
=
\mathbf{S}_t\bar{\mathbf{q}}_t,
\qquad
\mathbf{r}_t^m
=
\mathbf{S}_t\bar{\mathbf{k}}_t^m ,
\end{equation}
where $\mathbf{r}_t^q$ denotes the main query readout, while $\mathbf{r}_t^m$
probes the recurrent state along the memory-conditioned direction.
We then compute the query-memory relation score and the adaptive suppression
coefficient as:
\begin{equation}
a_t^r
=
\operatorname{ReLU}
\left(
(\bar{\mathbf{k}}_t^m)^{\top}
\bar{\mathbf{q}}_t
\right),
\qquad
\eta_t^r
=
\frac{\|\mathbf{r}_t^m\|_2}
{\|\mathbf{r}_t^q\|_2+\|\mathbf{r}_t^m\|_2+\epsilon},
\end{equation}
where $a_t^r$ measures whether the current query accesses the memory-conditioned
direction, and $\eta_t^r$ measures the relative strength of the memory-direction
response compared with the main query readout. Therefore, the final TDGD
readout is formulated as:
\begin{equation}
\tilde{\mathbf{r}}_t
=
\mathbf{r}_t^q
-
\eta_t^r a_t^r \mathbf{r}_t^m \\
=
\mathbf{S}_t
\left(
\mathbf{I}
-
{\color{red}
\eta_t^r a_t^r
\bar{\mathbf{k}}_t^m
(\bar{\mathbf{k}}_t^m)^{\top}}
\right)
\bar{\mathbf{q}}_t .
\end{equation}
This memory-guided differential readout suppresses responses along repeated memory-aligned semantics, allowing memory-unexplained transition signals to dominate the current prediction. 
Finally, TDGD then applies output normalization and projection to the corrected readout, followed by the block-level gate:
\begin{equation}
\mathbf{o}_t
=
\mathbf{u}_t
\odot
\operatorname{Post}
\left(
\tilde{\mathbf{r}}_t
\right),
\end{equation}
where $\operatorname{Post}(\cdot)$ denotes the output normalization and linear projection of the GDN layer. 
The gate $\mathbf{u}_t$ further modulates the corrected readout before the lightweight linear layer, preventing noisy or redundant state responses from being uniformly propagated.

\paragraph{Prediction and Optimization.}
SinkRec predicts the next item from the output representation $\mathbf{o}_t$ after stacked hybrid blocks:
$p(i_{t+1}=j \mid i_{1:t})=\operatorname{softmax}(\mathbf{o}_t\mathbf{E}^{\top})_j$,
where $\mathbf{E}$ is the item embedding matrix. 
For efficient autoregressive training, given the positive target $i_{t+1}$, we sample $N$ negative items $\mathcal{N}_t$ and optimize the sampled softmax loss~\cite{ye2026fuxi}.
The final objective is 
$\mathcal{L}=\mathcal{L}_{\mathrm{pred}}+\lambda_{\mathrm{vq}}\mathcal{L}_{\mathrm{mem}}$.

\subsection{Complexity Analysis}
\paragraph{Complexity Analysis.}
SinkRec preserves the linear complexity of Gated DeltaNet with lightweight memory overhead. 
Let $N$ be the sequence length, $d$ the hidden dimension, $H$ the number of heads, $W$ the window size, $L$ the number of residual VQ levels, and $K$ the codebook size. 
The Gated DeltaNet backbone costs $O(Nd^2+Nd^2/H)$. 
SinkRec adds two main costs: window encoding with $O(NWd)$ for contiguous and dilated windows, and residual VQ lookup with $O(NLKd)$ over $L$ codebooks of size $K$. 
The memory-guided write and read operations are of the same order as the recurrent state computation, i.e., $O(Nd^2/H)$, and do not change the asymptotic scaling. 
Thus, the overall complexity is $O(Nd^2+Nd^2/H+NWd+NLKd)$. 
Since $W$, $L$, and $K$ are fixed hyperparameters, SinkRec scales linearly with $N$, remaining more efficient than Transformer-based recommenders with $O(N^2d)$ attention cost.

\begin{table}[t]
 \caption{Dataset statistics.}
 \centering
 	\setlength{\tabcolsep}{1mm}
 \begin{tabular}{@{} c|c|c|c|c @{}}
 \hline
 \textbf{Dataset} 	  & \textbf{User}   & \textbf{Item} & \textbf{Interactions} & \textbf{Avg. Len.}   \\
 \hline
 MovieLens-20M & 138,493 & 26,744 & 20,000,263 & 144.41   \\
 KuaiRec & 7,176 & 9,958 & 12,529,113 & 1745.97 \\
 Industrial & 70 M & 900 K & 500 M & 2.3 K \\
 \hline
\end{tabular}
\label{tab:dataset_statistics}
\vspace{-6mm}
\end{table}

\section{Experiments}
\label{experiments}
\subsection{Models and Datasets}
\textbf{Models.} 
To comprehensively evaluate the performance of SinkRec, we consider three representative categories of long-sequence recommendation baselines: full-attention models, linear-attention models, and Memory models. Specifically, for full-attention methods, we select SASRec~\cite{kang2018self}, HSTU~\cite{zhai2024actions}, Longer~\cite{chai2025longer}, BlossomRec~\cite{ma2026blossomrec}, HyTRec~\cite{xin2026hytrec}, CollectiveKV~\cite{li2026collectivekv}, which model user behavior sequences through standard or enhanced attention mechanisms. For linear-attention methods, we compare with Mamba4Rec~\cite{liu2024mamba4rec}, TiM4Rec~\cite{fan2025tim4rec}, FuXi-Linear~\cite{ye2026fuxi}, which improve sequence modeling efficiency through recurrent, state-space, or linearized attention structures. For memory-enhanced methods, we include MSN~\cite{wu2026msn}, DUIA~\cite{liu2024dynamic}, which uses memory modules to store personalized user representations for long-term preference modeling.

\textbf{Datasets.}
To evaluate the effectiveness of SinkRec, we conduct experiments on two representative public datasets for long-sequence recommendation: MovieLens-20M and KuaiRec. MovieLens-20M\footnote{https://grouplens.org/datasets/movielens/} is a standard movie recommendation benchmark with approximately 20 million ratings and rich tagging information, suitable for evaluating long-term preference modeling. KuaiRec\footnote{https://kuairec.com/}~\cite{gao2022kuairec} is a public short-video recommendation dataset with diverse feedback signals and dynamic user behaviors, providing a challenging scenario for modeling rapidly evolving user interests. 
For the industrial dataset, we construct KuaiLLSR (i.e., industrial dataset) from large-scale behavioral logs on the Kuaishou platform, where users interact with local life service short videos. It contains eight days of uniformly sampled user-video interaction records and is used for model training and evaluation.

\subsection{Comparison with Baseline}
To validate the effectiveness of the proposed SinkRec, we report the overall performance of all baselines and SinkRec in Table~\ref{tab:preformance}. From the results, we obtain the following observations.
First, conventional Transformer-based methods show limited performance in long-sequence recommendation, indicating that relying solely on contextual interactions is insufficient for effectively modeling long user histories. In contrast, linear-attention-based models demonstrate strong potential, as they provide a more efficient and scalable way to capture long-range sequential dependencies.
Second, the effectiveness of time-aware methods such as FuXi-Linear shows that temporal periodicity and temporal bias are important for sequential prediction. As contextual biases, temporal signals can enhance the modeling of behavior dependencies in long sequences. Moreover, target-time-aware modeling further contributes to performance improvement by aligning historical behaviors with the prediction time.
Third, SinkRec consistently outperforms all baseline methods, demonstrating the effectiveness of decoupling static semantic retrieval from dynamic state computation. The results also verify that incorporating time-aware state modeling improves the ability of recurrent memory-based models to capture evolving user interests.

\begin{table}[t]
\setlength{\abovecaptionskip}{2pt}
\setlength{\belowcaptionskip}{2pt}
\centering
\caption{Overall performance comparison on public and industrial datasets. The best results are highlighted in bold, and the second-best results are underlined. The performance gains of our method are statistically significant under a paired $t$-test with $p < 0.01$.}
\label{tab:performance_comparison}
\setlength{\tabcolsep}{0.7mm}
\small
\resizebox{\linewidth}{!}{
\begin{tabular}{c|ccccc|ccccc|ccccc}
\hline
\textbf{Dataset} 
& \multicolumn{5}{c|}{\textbf{KuaiRec}} 
& \multicolumn{5}{c|}{\textbf{MovieLens-20M}} 
& \multicolumn{5}{c}{\textbf{Industrial}} \\
\hline
\textbf{Model} 
& R@10 & R@50 & N@10 & N@50 & MRR
& R@10 & R@50 & N@10 & N@50 & MRR
& R@10 & R@50 & N@10 & N@50 & MRR \\
\hline

\textbf{SASRec} 
& 0.1154 & 0.1612 & 0.1895 & 0.4027 & 0.1059
& 0.1787 & 0.2356 & 0.3138 & 0.5713 & 0.1524
& 0.05594 & 0.17404 & 0.03021 & 0.05336 & 0.02602 \\

\textbf{Mamba4Rec} 
& 0.1165 & 0.1622 & 0.1951 & 0.4068 & 0.1058
& 0.1862 & 0.2426 & 0.3220 & 0.5765 & 0.1595
& 0.05324 & 0.24153 & 0.02813 & 0.06168 & 0.02813 \\

\textbf{TiM4Rec} 
& 0.1149 & 0.1609 & 0.1894 & 0.4029 & 0.1053
& 0.1850 & 0.2418 & 0.3206 & 0.5772 & 0.1584
& 0.07297 & 0.22342 & 0.03379 & 0.06328 & 0.03185 \\

\textbf{HSTU} 
& 0.1265 & 0.1755 & 0.2074 & 0.4358 & 0.1155
& 0.2095 & 0.2663 & 0.3548 & 0.6107 & 0.1798
& 0.09135 & 0.29139 & 0.04056 & 0.07669 & 0.03834 \\

\textbf{LONGER} 
& 0.1122 & 0.1583 & 0.1873 & 0.4019 & 0.1023
& 0.1631 & 0.2206 & 0.2893 & 0.5497 & 0.1399
& 0.09572 & 0.30156 & 0.04656 & 0.08522 & 0.04665 \\

\textbf{HyTRec} 
& 0.1131 & 0.1566 & 0.1862 & 0.3878 & 0.1036
& 0.1814 & 0.2397 & 0.3168 & 0.5694 & 0.1547
& 0.09259 & 0.26657 & 0.04509 & 0.08135 & 0.04720 \\

\textbf{MSN} 
& 0.1340 & 0.1781 & 0.2214 & 0.4436 & 0.1208
& 0.2098 & 0.2673 & 0.3539 & 0.6133 & 0.1805
& 0.08968 & 0.25544 & 0.03367 & 0.08557 & 0.03930 \\

\textbf{BlossomRec} 
& 0.1085 & 0.1502 & 0.1812 & 0.3739 & 0.0988
& 0.1742 & 0.2315 & 0.3056 & 0.5578 & 0.1476
& 0.08430 & 0.25354 & 0.04264 & 0.07552 & 0.03792 \\

\textbf{DUIA} 
& \underline{0.1391} & \underline{0.1898} & \underline{0.2248} & \underline{0.4588} & \underline{0.1271}
& 0.2116 & 0.2694 & 0.3571 & 0.6152 & 0.1819
& 0.09084 & 0.26848 & 0.05140 & 0.07798 & 0.04658 \\

\textbf{CollectiveKV} 
& 0.1297 & 0.1790 & 0.2117 & 0.4405 & 0.1184
& \underline{0.2144} & \underline{0.2723} & \underline{0.3730} & 0.6113 & \underline{0.1832}
& \underline{0.10736} & 0.29970 & \underline{0.06048} & 0.08586 & \underline{0.04887} \\

\textbf{FuXi-Linear} 
& 0.1368 & 0.1851 & 0.2242 & 0.4486 & 0.1235
& 0.2131 & 0.2700 & 0.3592 & \underline{0.6161} & 0.1830
& 0.10372 & \underline{0.33193} & 0.04493 & \underline{0.08843} & 0.04208 \\

\hline
\textbf{SinkRec} 
& \textbf{0.1515} & \textbf{0.2022} & \textbf{0.2460} & \textbf{0.4805} & \textbf{0.1365}
& \textbf{0.2276} & \textbf{0.2867} & \textbf{0.3862} & \textbf{0.6437} & \textbf{0.1973}
& \textbf{0.12835} & \textbf{0.35781} & \textbf{0.07693} & \textbf{0.11247} & \textbf{0.05147} \\

\hline
\end{tabular}
}
\vspace{-12pt}
\label{tab:preformance}
\end{table}

\subsection{Ablation Study}\label{sec:ablation_study}
As shown in Table~\ref{tab:ablation_study}, we evaluate the contribution of each key component through four variants: 
(1) w/o TDGD, which replaces TDGD with vanilla Gated DeltaNet; 
(2) w/o Memory, which removes the conditional memory module and uses only TDGD; 
(3) w/o Time, which removes temporal decay and periodic temporal features; and 
(4) w/o Difference, which replaces the differential gate $\alpha$ with a content-only gate.

From the ablation results, we draw three main observations. 
First, TDGD improves performance by integrating temporal modeling with differential gating, enabling more effective state writing and reading. 
Second, the memory module captures reusable local behavioral patterns and guides TDGD to focus on memory-unexplained transitions. 
Third, temporal features and differential readout are both essential: temporal signals make state updates time-aware, while differential readout reduces memory-aligned responses and prevents recurrent states from repeatedly absorbing redundant semantic patterns.

\begin{table}[htbp]
    \centering
    \vspace{-12pt}
    \caption{Ablation Study of SinkRec on Kuairec.}
    \label{tab:ablation_study}
    \setlength{\tabcolsep}{0.8mm}{
    \small
    \begin{tabular}{lccccc}
        \toprule
        \textbf{Setting} & \textbf{R@10} & \textbf{R@50} & \textbf{N@10} & \textbf{N@50} & \textbf{MRR} \\
        \midrule
        Base                  & \textbf{0.1515} & \textbf{0.2022} & \textbf{0.2460} & \textbf{0.4805} & \textbf{0.1365} \\
        1) w/o TDGD & 0.1198 & 0.1686 & 0.2062 & 0.4327 & 0.1071 \\
        2) w/o Memory & 0.1405 & 0.1872 & 0.2275 & 0.4569 & 0.1276 \\
        3) w/o time & 0.1236 & 0.1731 & 0.2092 & 0.4362 & 0.1113 \\
        4) w/o difference & 0.1455 & 0.1942 & 0.2345 & 0.4649 & 0.1321 \\
        \bottomrule
    \end{tabular}
    }
    \vspace{-12pt}
\end{table}

\begin{wrapfigure}[11]{r}{0.56\linewidth}
    \centering
    \vspace{-10pt}
    \includegraphics[width=\linewidth]{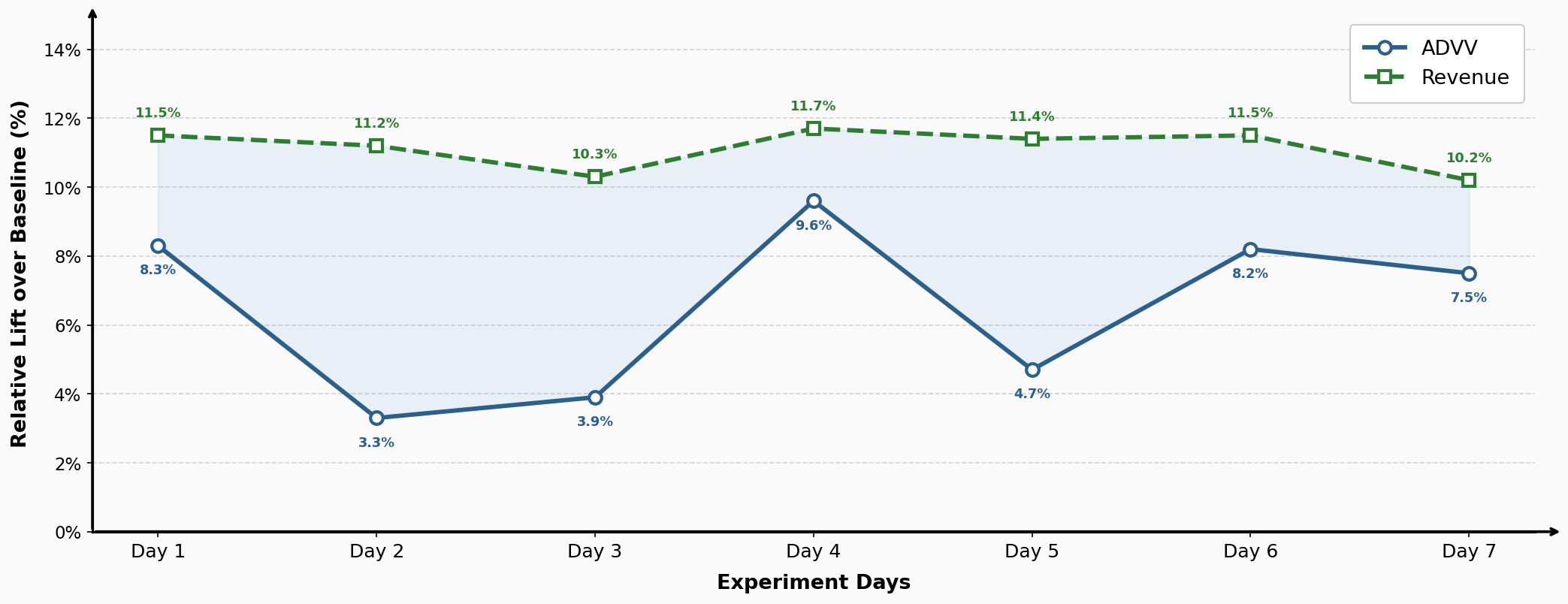}
    \vspace{-8pt}
    \caption{Online A/B experimental results.}
    \label{fig:gdla}
    \vspace{-12pt}
\end{wrapfigure}

\subsection{Online A/B Experimental Results}
To evaluate the practical effectiveness of SinkRec, we conducted a 7-day online A/B test on Kuaishou’s advertising platform with live production traffic. The control and treatment groups are each allocated 10\% of the total traffic for a fair comparison.

As shown in Figure~\ref{fig:gdla}, our method achieves consistent improvements over the control group, with average relative gains of 6.5\% in ADVV and 11.1\% in Revenue. Both improvements are statistically significant under a two-tailed test (p < 0.01). The gains remain positive throughout the experimental period, demonstrating that the proposed method can effectively improve both advertiser-side value and platform revenue in production environments. These results validate the practical effectiveness, robustness, and scalability of our framework in large-scale industrial advertising recommendation systems.

\section{Conclusion}
We present SinkRec, an efficient long-sequence recommendation framework that addresses the semantic state sink phenomenon in recurrent linear attention. By identifying that recurring semantic interests can be repeatedly absorbed into the recurrent state and interfere with current interest adaptation, we motivate a memory-transition decoupled design that separates collaborative semantic lookup from dynamic transition computation. 
SinkRec externalizes recurring interests into Conditional Semantic Memory and introduces Time-aware Differential Gated DeltaNet to separate sink-inducing collaborative semantics from recurrent state dynamics. This reduces semantic dominance in recurrent states and improves sensitivity to recent interest shifts and target-relevant behaviors. Experiments on public and industrial datasets demonstrate that SinkRec improves recommendation effectiveness while preserving the efficiency of recurrent linear attention.

{
\small
\bibliographystyle{plainnat}
\bibliography{neurips_2026}
}


\newpage
\appendix
\section{Theoretical Analysis of Semantic State Sink}
\label{app:recurrent_attention_sink}

We provide a simplified derivation to explain why recurrent linear attention may suffer from semantic state sink in long-sequence recommendation. 
Consider the Gated DeltaNet update:
\begin{equation}
    \mathbf{S}_t
    =
    \alpha_t \mathbf{S}_{t-1}
    \left(
    \mathbf{I}
    -
    \beta_t\mathbf{k}_t\mathbf{k}_t^\top
    \right)
    +
    \beta_t\mathbf{v}_t\mathbf{k}_t^\top ,
    \label{eq:app_gdn_update}
\end{equation}
where $\mathbf{S}_t$ is the recurrent state, $\mathbf{k}_t$ and $\mathbf{v}_t$ are the key and value vectors, and $\alpha_t,\beta_t$ control state retention and writing strength. 
This update writes the value $\mathbf{v}_t$ into the state subspace addressed by $\mathbf{k}_t$.
In long user sequences, recurring semantic behavior patterns often appear across distant positions and are expected to produce similar key representations. 
To analyze how such recurrence affects a key-addressed state subspace, we consider an idealized setting where these keys are aligned with a normalized direction $\mathbf{u}$:
\begin{equation}
    \mathbf{k}_t=\mathbf{u}, 
    \qquad 
    \|\mathbf{u}\|_2=1 .
\end{equation}
Let $\mathbf{z}_t=\mathbf{S}_t\mathbf{u}$ denote the state content stored in this repeatedly accessed direction. 
Right-multiplying Eq.~\eqref{eq:app_gdn_update} by $\mathbf{u}$ and using $\mathbf{u}^{\top}\mathbf{u}=1$ yields:
\begin{equation}
\begin{aligned}
    \mathbf{z}_t
    &=
    \alpha_t(1-\beta_t)\mathbf{z}_{t-1}
    +
    \beta_t\mathbf{v}_t .
\end{aligned}
\label{eq:app_address_update}
\end{equation}
Thus, the repeatedly accessed semantic direction follows a first-order state recurrence.
To make the address-level dynamics analytically tractable, we consider a locally stationary setting where the retention and write strengths are time-invariant, i.e., $\alpha_t=\alpha$ and $\beta_t=\beta$. 
We define the effective retention factor as $\rho=\alpha(1-\beta)$, with $0\leq\rho<1$. 
Under this approximation, Eq.~\eqref{eq:app_address_update} reduces to:
\begin{equation}
    \mathbf{z}_t
    =
    \rho\mathbf{z}_{t-1}
    +
    \beta\mathbf{v}_t .
\end{equation}
This recurrence characterizes the evolution of the content stored in a repeatedly accessed semantic address: previous address-specific content is retained by $\rho$, while the current value is written with strength $\beta$.
For semantically recurrent behaviors, the written values usually share a stable semantic component. 
We therefore decompose each value vector as:
\begin{equation}
    \mathbf{v}_t
    =
    \bar{\mathbf{v}}
    +
    \boldsymbol{\epsilon}_t ,
\end{equation}
where $\bar{\mathbf{v}}$ denotes the recurring semantic component and $\boldsymbol{\epsilon}_t$ denotes instance-specific residual variation.
Combining the recurrence with the semantic decomposition of $\mathbf{v}_t$, we can express the address-specific state as:
\begin{equation}
\mathbf{z}_t
=
\rho^t\mathbf{z}_0
+
\underbrace{
\frac{\beta(1-\rho^t)}{1-\rho}\bar{\mathbf{v}}
}_{\text{accumulated recurring semantics}}
+
\underbrace{
\beta\sum_{i=1}^{t}\rho^{t-i}\boldsymbol{\epsilon}_i
}_{\text{residual variations}} .
\label{eq:app_sink_decomposition}
\end{equation}
This decomposition separates the address-specific state into a decayed initialization term, an accumulated recurring semantic term, and a residual variation term. 
Since $\rho<1$, the state does not diverge; instead, as the same semantic address is repeatedly visited, the contribution of $\bar{\mathbf{v}}$ approaches $\frac{\beta}{1-\rho}\bar{\mathbf{v}}$, causing this address subspace to become biased toward the recurring semantic pattern.
Such an address-level bias becomes harmful when the next behavior at time $t+1$ still activates the same semantic address $\mathbf{u}$, while its value carries an additional residual transition component:
\begin{equation}
    \mathbf{v}_{t+1}
    =
    \bar{\mathbf{v}}
    +
    \mathbf{r}_{t+1},
\end{equation}
where $\mathbf{r}_{t+1}$ denotes the new transition component. 
Ignoring the residual variation and using the steady-state approximation 
\(\mathbf{z}_t \approx \frac{\beta}{1-\rho}\bar{\mathbf{v}}\), the next state content becomes:
\begin{equation}
\begin{aligned}
\mathbf{z}_{t+1}
&=
\rho\mathbf{z}_t
+
\beta(\bar{\mathbf{v}}+\mathbf{r}_{t+1}) \\
&=
\underbrace{
\frac{\beta}{1-\rho}\bar{\mathbf{v}}
}_{\text{historical recurring semantics}}
+
\underbrace{
\beta\mathbf{r}_{t+1}
}_{\text{new transition signal}} .
\end{aligned}
\label{eq:app_transition_mixture}
\end{equation}
Therefore, the new transition signal is mixed with an accumulated historical semantic component. 
Their relative strength can be approximated by:
\begin{equation}
    \mathcal{D}_{t+1}
    \approx
    \frac{
    \left\|\frac{\beta}{1-\rho}\bar{\mathbf{v}}\right\|_2
    }{
    \left\|\beta\mathbf{r}_{t+1}\right\|_2
    }
    =
    \frac{
    \|\bar{\mathbf{v}}\|_2
    }{
    (1-\rho)\|\mathbf{r}_{t+1}\|_2
    } .
    \label{eq:app_sink_ratio}
\end{equation}
When the effective retention $\rho$ is large, or when the new transition signal $\mathbf{r}_{t+1}$ is weak, the accumulated recurring semantics dominate the address-specific state content.

To connect this address-level effect back to the Gated DeltaNet output, let
\(\mathbf{o}_{t+1}\) denote the state readout. 
For a query \(\mathbf{q}_{t+1}\) with a non-negligible projection on the same semantic direction,
\begin{equation}
    \mathbf{q}_{t+1}
    =
    c_{t+1}\mathbf{u}
    +
    \mathbf{q}_{t+1}^{\perp},
    \qquad
    c_{t+1}
    =
    \mathbf{u}^{\top}\mathbf{q}_{t+1}.
\end{equation}
The readout can then be decomposed as:
\begin{equation}
    \mathbf{o}_{t+1}
    =
    \mathbf{S}_{t+1}\mathbf{q}_{t+1}
    =
    c_{t+1}\mathbf{z}_{t+1}
    +
    \mathbf{S}_{t+1}\mathbf{q}_{t+1}^{\perp}.
    \label{eq:app_query_readout}
\end{equation}
Thus, when \(c_{t+1}\) is non-negligible, the accumulated recurring semantic component in \(\mathbf{z}_{t+1}\) is directly propagated to the output \(\mathbf{o}_{t+1}\). 
A large \(\mathcal{D}_{t+1}\) therefore indicates that the readout along this semantic direction is dominated by historical recurring semantics, making the new transition signal relatively less salient.

This phenomenon indicates that semantic state sink is not characterized by unbounded growth of the state norm. 
Rather, recurrent updates induced by repeated behaviors gradually concentrate the state along frequently visited key-addressed semantic subspaces. 
Consequently, in long-sequence recommendation, the recurrent state becomes biased toward dominant historical patterns, reducing its sensitivity to recent preference shifts and sparse but predictive transition signals.

\section{Comparison with Vanilla Gated DeltaNet}
\label{app:tdgd_vs_gdn}
Table~\ref{tab:gdn_tdgd_compare} summarizes the architectural differences between vanilla Gated DeltaNet and TDGD. 
Vanilla Gated DeltaNet updates a compact recurrent state by writing the current key--value pattern and then reads the state with the current query. 
This design is efficient, but its writing and reading operations are both driven only by the current input representation. 
Therefore, when similar semantic behaviors repeatedly appear in a long sequence, vanilla Gated DeltaNet may continue to write these recurring patterns into the state and later retrieve them through query-aligned directions, which can cause semantic state sinks.

TDGD differs from vanilla Gated DeltaNet in two key aspects. 
First, on the writing side, TDGD replaces the raw value $\mathbf{v}_t$ with the memory-filtered value $\bar{\mathbf{v}}_t$ and uses the time-aware key $\tilde{\mathbf{k}}_t$. 
This makes state updates aware of whether the current behavior is already covered by retrieved semantic memory, thereby reducing redundant writes of recurring interests. 
Second, on the reading side, TDGD augments the standard state readout $\mathbf{S}_t\mathbf{q}_t$ with a state-relation differential term 
$\eta_t^r a_t^r\mathbf{S}_t\bar{\mathbf{k}}_t^m$. 
This term suppresses the memory-key-aligned response from the same recurrent state, preventing memory-dominated semantic directions from overwhelming the final readout.

These two modifications target both the formation and the effect of semantic state sinks. 
The memory-conditioned write gate reduces redundant semantic accumulation before it enters the recurrent state, while the state-relation differential readout suppresses memory-aligned responses already encoded in the state. 
This enables TDGD to preserve memory-unexplained transition signals, making the model more responsive to recent interest shifts and infrequent but currently relevant behaviors in long sequence recommendation.

\begin{table}[t]
\centering
\caption{Comparison between vanilla Gated DeltaNet and TDGD.}
\label{tab:gdn_tdgd_compare}
\resizebox{\columnwidth}{!}{
\begin{tabular}{lcc}
\toprule
\textbf{Component} 
& \textbf{Vanilla Gated DeltaNet} 
& \textbf{TDGD} \\
\midrule

State writing 
& 
$\beta_t(\mathbf{v}_t-\alpha_t\mathbf{S}_{t-1}\mathbf{k}_t)\mathbf{k}_t^\top$
& 
$\beta_t(\bar{\mathbf{v}}_t-\alpha_t\mathbf{S}_{t-1}\tilde{\mathbf{k}}_t)\tilde{\mathbf{k}}_t^\top$ \\

Write signal
&
Raw input value $\mathbf{v}_t$
&
Memory-filtered value $\bar{\mathbf{v}}_t$ \\

Key representation
&
Content key $\mathbf{k}_t$
&
Time-aware key $\tilde{\mathbf{k}}_t$ \\

Memory awareness 
& 
No explicit memory conditioning
& 
Memory-conditioned write gate \\

State readout 
& 
$\mathbf{r}_t=\mathbf{S}_t\mathbf{q}_t$
& 
$\tilde{\mathbf{r}}_t
=
\mathbf{S}_t\bar{\mathbf{q}}_t
-
\eta_t^r a_t^r\mathbf{S}_t\bar{\mathbf{k}}_t^m$ \\

Readout control
&
Query-only state access
&
Relation-level memory suppression \\

Sink control 
& 
Implicit through decay and gating
& 
Explicit write suppression and differential readout \\

Targeted effect
&
General recurrent filtering
&
Mitigates repetitive semantic state sink \\

Complexity 
& 
Linear recurrent computation
& 
Linear recurrent computation with lightweight memory conditioning \\

\bottomrule
\end{tabular}
}
\end{table}

\section{Illustration of Historical Influence Score}
\label{app:state_influence}

To visualize semantic state sink, we compute a historical influence score that measures how strongly each past behavior affects the current prediction through the recurrent state. 
For model $f\in\{\mathrm{base},\mathrm{ours}\}$, the state update at position $t$ is:
\begin{equation}
\Delta \mathbf{S}_t^{f}
=
\beta_t^{f}
\left(
\mathbf{v}_t^{f}
-
\bar{\mathbf{S}}_{t-1}^{f}\mathbf{k}_t^{f}
\right)
(\mathbf{k}_t^{f})^{\top},
\qquad
\bar{\mathbf{S}}_{t-1}^{f}
=
\alpha_t^{f}\mathbf{S}_{t-1}^{f}.
\end{equation}
Given the current prediction position $T$, we define:
\begin{equation}
I_t^{f}
=
\left\|
\Gamma_{t\rightarrow T}^{f}
\Delta \mathbf{S}_t^{f}
\mathbf{q}_{T}^{f}
\right\|_2 ,
\end{equation}
where $\Gamma_{t\rightarrow T}^{f}$ denotes the accumulated retention from $t$ to $T$, and $\mathbf{q}_{T}^{f}$ is the current readout query. 
Thus, $I_t^{f}$ measures the readable influence of the $t$-th historical behavior on the current prediction.

For the case visualization in Figure~\ref{fig:intro}, we rescale the scores with a shared maximum over the compared models:
\begin{equation}
\tilde{I}_t^{f}
=
\frac{
I_t^{f}
}{
\max\limits_{g\in\{\mathrm{base},\mathrm{ours}\},\,i}
I_i^{g}
+\epsilon
}.
\end{equation}
This shared scaling preserves the relative magnitude between the vanilla model and SinkRec. 
Higher values indicate that the corresponding historical behavior occupies a stronger role in the current recurrent readout. 
Semantic state sink is indicated when semantically repetitive but target-irrelevant behaviors obtain disproportionately high influence scores and dominate the prediction.

\section{Experiment Settings, Metrics, and Model Scale}
\textbf{Experiment settings.} For data preprocessing, we follow the same settings as HSTU and FuXi-Linear for all compared models to ensure fair comparison. Beyond public datasets, we further validate our framework on a large-scale industrial dataset derived from real-world impression logs of Kuaishou’s advertising platform. Compared with standard public benchmarks, this dataset provides a more realistic evaluation environment characterized by large-scale traffic, diverse user behaviors, and practical commercial recommendation constraints. The results demonstrate the scalability and robustness of our method in real-world industrial deployment scenarios. 

Training is conducted on 2 A800 GPUs, with all settings following Fuxi-Linear. Instead of stacking distinct layers, SinkRec iteratively reuses a shared hybrid memory-transition block, resulting in a substantially smaller parameter budget.

\textbf{Metrics.}
We adopt three widely used ranking metrics for evaluation: top-$K$ Hit Ratio (HR@$K$, denoted as R@$K$ in tables), Normalized Discounted Cumulative Gain (NDCG@$K$, denoted as N@$K$ in tables), and Mean Reciprocal Rank (MRR). Higher values indicate better recommendation performance for all metrics. Unless otherwise specified, we rank the ground-truth item against the full item set and report results at $K=10$ and $K=50$. For the online A/B experiments, we use Advertiser Value (ADVV) and Revenue as core evaluation metrics.

\textbf{Model scale.} For KuaiRec and ML-20M, we set the number of layers to 4 and 8, and the embedding dimension to 128 and 256, respectively. To ensure a fair comparison, we tune the hidden dimensions of all baselines to keep their model scales comparable. The detailed hyperparameter settings and the numbers of non-embedding parameters are reported in Table~\ref{tab:model_size}. These results also show that SinkRec achieves competitive performance with a more lightweight architecture.

It is worth noting that SinkRec shares parameters across hybrid blocks, which substantially reduces the overall parameter count without increasing inference latency.

\begin{table}[htbp]
\centering
\caption{The sizes of models across various datasets.}
 \setlength{\tabcolsep}{1mm}{
    \small
    \begin{tabular}{lcccc}
    \toprule
    \textbf{Model} & \textbf{Time-aware} & \textbf{ML-20M} & \textbf{KuaiRec} \\
    \midrule
    SASRec (4d)    & $\times$ & 6.31M & 791.04K \\
    Mamba4Rec (4d)      & $\times$ & 7.91M & 1.04M \\
    TiM4Rec (1d)        & $\checkmark$ & 7.57M & 984.16K \\
    HSTU (4d)      & $\checkmark$ & 7.40M & 926.81K \\
    FuXi-Linear (1d)      & $\checkmark$ & 7.34M & 917.66K \\ \hline
    SinkRec (4d)      & $\checkmark$ & 4.52M & 541.32K \\
    \bottomrule
    \end{tabular}
}
\label{tab:model_size}
\end{table}

\begin{figure*}[!t]
\centering	
\includegraphics[width=\textwidth]{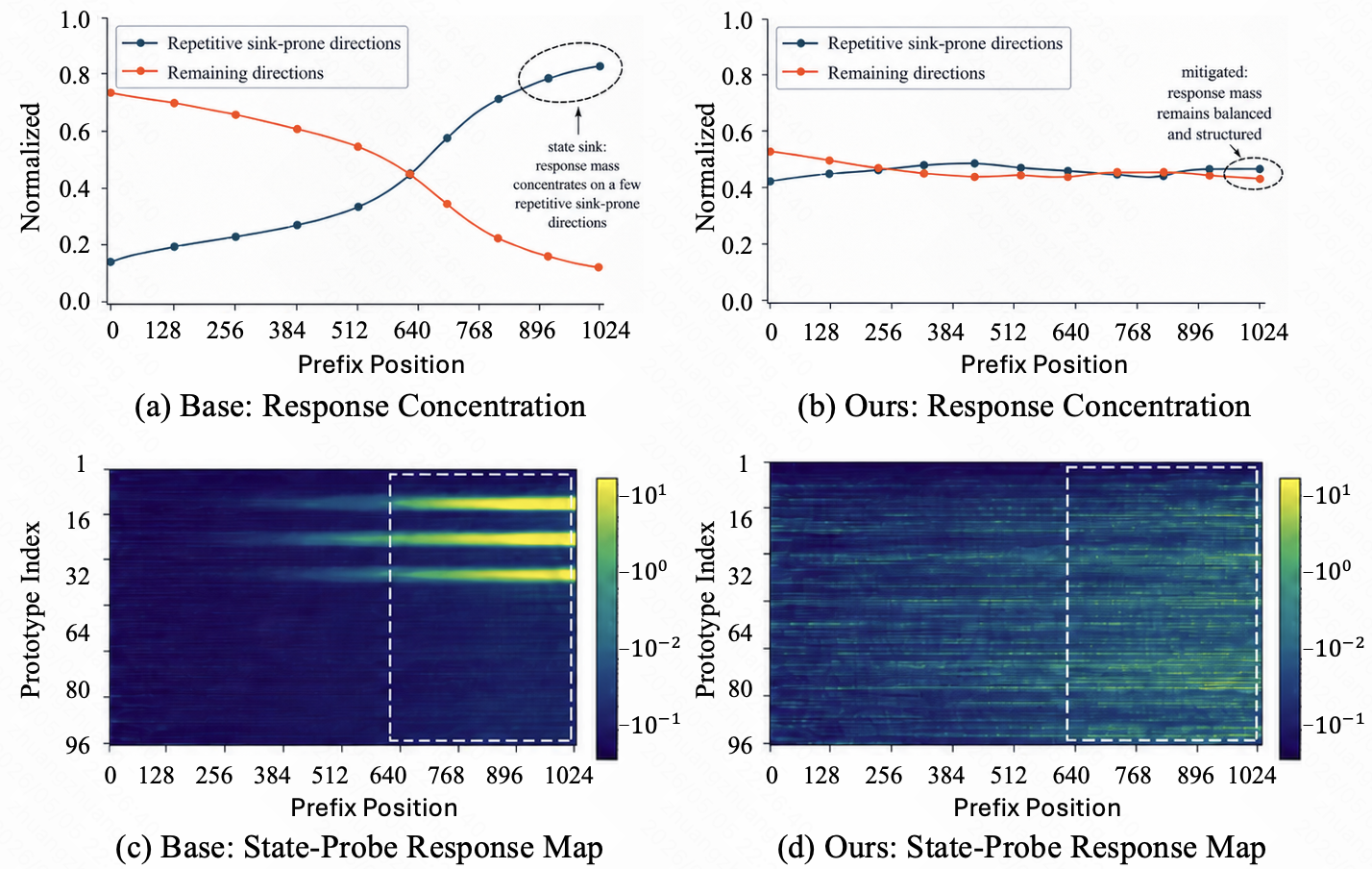}
\caption{
\textbf{TDGD alleviates semantic state sink.}
(a) In the base model, normalized state-response mass increasingly concentrates on repetitive sink-prone directions as the prefix grows.
(b) TDGD keeps the response mass more balanced across repetitive and remaining directions.
(c) The base state-probe response map $\|\mathbf{S}_t\mathbf{k}_j^m\|_2$ shows persistent bright bands, indicating that a few repetitive prototypes dominate the recurrent state.
(d) TDGD suppresses such directional concentration while preserving structured semantic responses.
For (c) and (d), prototype indices are sorted by repetition frequency from high to low, with $j=1$ and $j=96$ corresponding to the most and least frequent prototypes, respectively.
}
\label{fig:state_sink_visualization}
\end{figure*}

\section{Mechanism Analysis: From Semantic Explanation to State Innovation}
\paragraph{Visualization of semantic state sink.}
Figure~\ref{fig:state_sink_visualization} analyzes semantic state sink by probing recurrent states with a shared semantic basis. 
Specifically, we use the learned memory codewords as semantic probe directions and apply the same probes to both the base model and SinkRec. 
Although the base model does not contain the memory module, its recurrent state can still be projected onto these codeword directions to examine whether it concentrates on repetitive semantics.

Given the recurrent state $\mathbf{S}_t$ and the projected memory-key direction $\mathbf{k}_j^m$ associated with the $j$-th codeword, we define the state-probe response as
\begin{equation}
e_{t,j}
=
\left\|
\mathbf{S}_t\mathbf{k}_j^m
\right\|_2 .
\end{equation}
A larger $e_{t,j}$ indicates that the recurrent state is more strongly readable along the semantic direction represented by the $j$-th codeword.

Let $\mathcal{J}_{\mathrm{rep}}$ denote the top-$K$ codeword directions with the highest usage frequency, which represent repetitive behavioral semantics. 
We measure the response-energy mass concentrated on these repetitive directions as
\begin{equation}
m_t^{\mathrm{rep}}
=
\frac{
\sum_{j\in\mathcal{J}_{\mathrm{rep}}} e_{t,j}^{2}
}{
\sum_{j=1}^{J} e_{t,j}^{2}+\epsilon
},
\qquad
m_t^{\mathrm{rem}}
=
1-m_t^{\mathrm{rep}} ,
\end{equation}
where $m_t^{\mathrm{rep}}$ quantifies how much state-probe energy is assigned to frequently reused semantic directions.

Figure~\ref{fig:state_sink_visualization}(a) shows that the base model gradually shifts more response mass toward high-frequency codeword directions as the prefix grows. 
This indicates that its recurrent state becomes increasingly concentrated on repetitive semantics. 
Figure~\ref{fig:state_sink_visualization}(b) shows that TDGD reduces this excessive concentration and maintains a more balanced response distribution, suggesting that memory-guided writing and reading alleviate semantic state sink.

Figures~\ref{fig:state_sink_visualization}(c) and (d) further visualize the full state-probe response map $\|\mathbf{S}_t\mathbf{k}_j^m\|_2$, where codeword directions are sorted by usage frequency. 
In the base model, a few high-frequency directions form persistent bright horizontal bands, especially at later prefixes, showing that repetitive semantics dominate the recurrent state response. 
In contrast, TDGD yields more distributed yet still structured responses across semantic directions, suppressing excessive dominance by repetitive patterns while preserving meaningful semantic organization.




\begin{figure*}[!t]
\centering	
\includegraphics[width=0.6\linewidth]{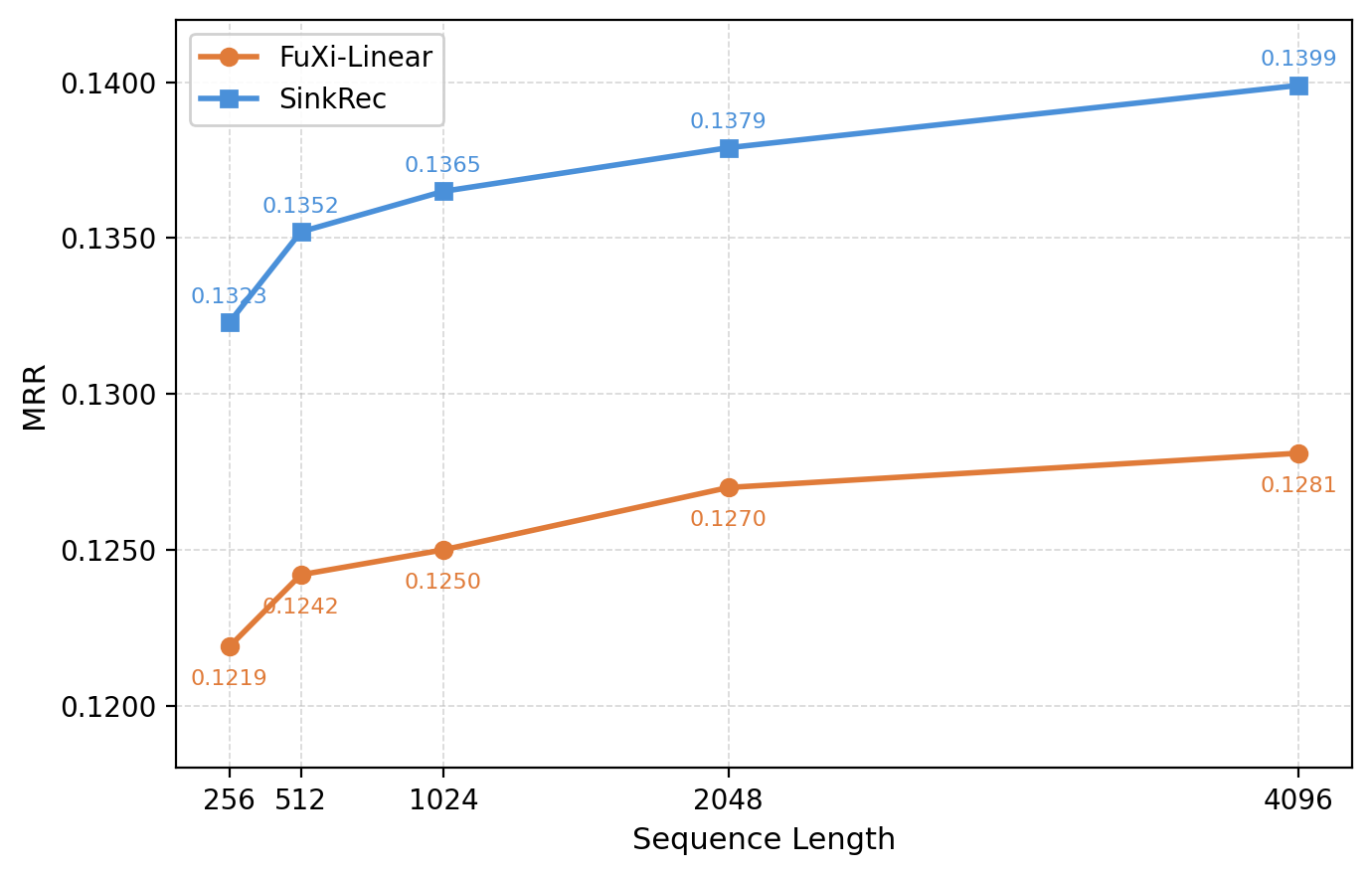}
\caption{The effectiveness comparison across different sequence lengths.}
\label{fig:scaling_up}
\end{figure*}
\section{Sequence Length Scaling Experiment}
To evaluate the scalability of SinkRec with respect to sequence length, we conduct comparative experiments on the KuaiRec dataset. 
We use FuXi-Linear, the strongest baseline, as the reference model and follow its reported hyperparameter settings for fair comparison. 
We gradually increase the input sequence length from 256 to 4096. 
As shown in Figure~\ref{fig:scaling_up}, both models benefit from longer histories, but SinkRec consistently outperforms FuXi-Linear across all sequence lengths. 
Specifically, SinkRec achieves relative MRR improvements of 8.53\%, 8.86\%, 9.20\%, 8.58\%, and 9.21\% at sequence lengths 256, 512, 1024, 2048, and 4096, respectively. 
Moreover, the performance gap remains stable and even reaches the largest relative gain at length 4096, indicating that SinkRec scales more effectively to longer histories. 
This demonstrates that the proposed memory-transition decoupled design can better exploit extended user sequences while maintaining robust architectural scalability.

\section{Computational Cost}
To further assess the computational efficiency of the proposed architecture, we report the per-batch inference time of all compared models under identical hardware settings. For a fair comparison of the core sequential modeling component, we remove the external Memory module from our method in this efficiency evaluation and measure the inference latency of the remaining backbone. As shown in Table~\ref{tab:infer_time}, our method achieves the fastest inference speed, requiring only 44.578 ms per batch. This is substantially lower than CollectiveKV (93.578 ms), HSTU (80.491 ms), and BlossomRec (80.368 ms). Notably, our method also outperforms the lightweight linear baseline FuXi-Linear, reducing the inference latency from 61.263 ms to 44.578 ms.

\begin{table}[t]
\centering
\setlength{\tabcolsep}{12pt}
\small
\caption{Inference time comparison on the same NVIDIA A800 GPUs.}
\label{tab:infer_time}
\begin{tabular}{lc}
\toprule
\textbf{Method} & \textbf{Time (ms/batch)} \\
\midrule
HSTU          & 80.491 \\
HyTRec        & 64.140 \\
MSN           & 73.543 \\
BlossomRec    & 80.368 \\
DUIA          & 72.052 \\
CollectiveKV  & 93.578 \\
FuXi-Linear   & 61.263 \\
\midrule
\textbf{SinkRec} & \textbf{44.578} \\
\bottomrule
\end{tabular}
\end{table}

The relatively high cost of CollectiveKV mainly comes from its cross-user collective attention mechanism, while HSTU and BlossomRec introduce additional computation through hierarchical modeling or multi-granularity feature interactions. In contrast, after removing the external Memory module, our backbone still maintains an efficient recurrent/linear sequential modeling structure and avoids expensive quadratic attention or cross-user aggregation during inference. These results demonstrate that the core design of our method is computationally efficient and suitable for large-scale long-sequence recommendation scenarios with strict online latency constraints.

\section{Discussions}
\paragraph{Semantic state sink in recurrent linear attention.}
Recurrent linear attention improves efficiency by compressing long histories into recurrent states. However, the same state must preserve collaborative semantic patterns and compute dynamic transitions. In long user histories, semantically recurrent local patterns can be repeatedly written into the state, causing high-frequency interests to dominate the representation. This \textit{semantic state sink} weakens the model's sensitivity to current local behaviors and explains why longer histories do not always yield proportional gains.

\paragraph{A unified perspective on memory-transition decoupling.}
SinkRec views long-sequence recommendation as a memory-transition decoupling problem. Existing methods often couple semantic pattern retrieval and transition modeling within attention weights, retrieved subsequences, memory slots, or recurrent states. In contrast, SinkRec assigns collaborative semantic patterns to Conditional Semantic Memory and leaves memory-unexplained transition residuals to the recurrent state. This separation reduces redundant semantic accumulation and preserves state capacity for dynamic interest adaptation.

\paragraph{Semantic initialization of memory codebooks.}
One future direction is to enrich the memory codebooks with item-side semantic knowledge. 
While the current residual VQ codebooks are learned end-to-end from behavior windows and mainly capture collaborative recurring patterns, recommendation items often contain category, tag, title, or description information. 
These textual attributes can be encoded by pretrained LLM embedding models to initialize semantic codebooks. 
By freezing the initialized codebook vectors during recommendation training, the memory module can preserve a stable external semantic space and avoid being fully collapsed into task-specific ID correlations. 
The trainable window encoder and projection adapters can then learn to retrieve and use these fixed semantic anchors, while the recurrent state focuses on memory-unexplained transitions. 
Such frozen semantic codebooks may better capture the latent intent behind contiguous and dilated behavior windows, improving interpretability, cold-start robustness, and generalization to sparse behavioral patterns.

\end{document}